\documentclass[conference]{IEEEtran}
\IEEEoverridecommandlockouts
\usepackage{cite}
\usepackage{amsmath,amssymb,amsfonts}
\usepackage{algorithmic}
\usepackage{graphicx}
\usepackage{textcomp}
\usepackage{xcolor}
\usepackage{array}
\usepackage{booktabs}
\usepackage{subcaption}
\usepackage{tikz}
\usetikzlibrary{plotmarks}
\usepackage{pgfplots}
\usepackage{cleveref}
\usepackage[absolute,overlay]{textpos}
\usepackage{cite}
 \pgfplotsset{compat=1.17}
\def\BibTeX{{\rm B\kern-.05em{\sc i\kern-.025em b}\kern-.08em
    T\kern-.1667em\lower.7ex\hbox{E}\kern-.125emX}}

\usepackage[normalem]{ulem}

\begin{document}

\title{
Graph Neural Networks for Parkinson's Disease Detection \\
\thanks{This work was supported by the Swiss National Science
Foundation project CRSII5\_202228 on “Characterisation of
motor speech disorders and processes".}
}

\author{
\thanks{\textsuperscript{*}Equal contribution}
\IEEEauthorblockN{Shakeel A.~Sheikh\textsuperscript{1*}, Yacouba Kaloga\textsuperscript{1*}, Md. Sahidullah\textsuperscript{2}, Ina Kodrasi\textsuperscript{1}}
\IEEEauthorblockA{
\textit{\textsuperscript{1}Idiap Research Institute, Martigny, Switzerland} \\ \textit{\textsuperscript{2}Institute for Advancing Intelligence, CG CREST, India} \\ 
} 

}

\maketitle

\begin{abstract}
Despite the promising performance of state-of-the-art approaches for Parkinson’s Disease (PD) detection, these approaches often analyze individual speech segments in isolation, which can lead to sub-optimal results. 
Dysarthric cues that characterize speech impairments from PD patients are expected to be related across segments from different speakers. Isolated segment analysis fails to exploit these inter-segment relationships. Additionally, not all speech segments from PD patients exhibit clear dysarthric symptoms, introducing label noise that can negatively affect the performance and generalizability of current approaches. To address these challenges, we propose a novel PD detection framework utilizing Graph Convolutional Networks (GCNs). By representing speech segments as nodes and capturing the similarity between segments through edges, our GCN model facilitates the aggregation of dysarthric cues across the graph, effectively exploiting segment relationships and mitigating the impact of label noise. Experimental results demonstrate the advantages of the proposed GCN model for PD detection and provide insights into its underlying mechanisms.

\end{abstract}


\section{Introduction}
\label{sec:intro}
Speech production, a highly intricate phenomenon, necessitates a series of complex processes involving the coordinated function of various articulators~\cite{kroger2019neural}. 
Each process is critical and relies on the seamless integration of neural, muscular, and auditory systems to produce spoken language~\cite{kroger2019neural}. These processes are influenced by factors such as neurological impairments, emotional state, context, and prior knowledge, highlighting the complexity of speech production~\cite{freed2023motor}. In particular, neurological disorders such as Parkinson's disease (PD) affect the control of the muscles involved in speech production, leading to speech dysarthria characterized by imprecise articulation, abnormal speech rhythm, breathiness, and hypernasality~\cite{stewart1995speech, darley1969differential, baghai2012automatic}.

\par 
To detect speech dysarthria, clinicians commonly use audio-perceptual tests to evaluate and analyze the speech patterns of patients~\cite{parvanehthesis}. However, these manual diagnostic tests are often labor-intensive, time-consuming, and subjective~\cite{pernon2022perceptual}. 
To address these challenges, various automatic methods for PD detection have been proposed, which can be generally grouped into two main categories, i.e., classical machine learning~\cite{kodrasi2020automatic, kodrasi2020spectro, vaiciukynas_detecting_2017, sheikh2024impact, wang_towards_2016, karan_investigation_2022} and deep learning approaches~\cite{janbakhshi2022adversarial, janbakhshi_stft, janbakhshi_ua, javanmardi2024pre, vasquez2017convolutional}. 
These approaches leverage a variety of speech representations, such as e.g., the short-time Fourier transform (STFT)~\cite{vasquezcorrea17_interspeech}, Mel frequency cepstral coefficients~\cite{hawi2022automatic}, Mel spectrograms~\cite{sheikh2024impact}, or self-supervised embeddings like wav2vec2 (w2v2)~\cite{javanmardi2023wav2vec, janbakhshi_ua}, integrated with diverse classifiers and architectures to achieve pathological speech detection.
Despite the promising performance of state-of-the-art automatic PD detection approaches, they generally focus on analyzing individual speech segments in isolation.
Such individual analysis are sub-optimal for two reasons. First, the manifestation of dysarthric cues - such as imprecise articulation, abnormal speech rhythm, breathiness, or hypernasality - is expected to be closely related across representations of different speech segments from various PD speakers. 
By analyzing speech segments in isolation, state-of-the-art PD detection approaches fail to take advantage of this relationship.
Second, not all speech segments from PD patients necessarily contain dysarthric cues, resulting in label noise present in the training data.
By analyzing speech segments in isolation, such label noise may considerably affect the accuracy and generalization of state-of-the-art PD detection approaches.

{\textcolor{black}{
To tackle these challenges, in this paper we propose using graph convolutional networks (GCNs)~\cite{Kipf16} for PD detection. 
Specifically, we construct a graph where nodes represent speech segments and edges reflect the relationships between these segments based on their similarity in input representations. 
Given their proven effectiveness for pathological speech detection in~\cite{javanmardi2023wav2vec, janbakhshi_ua, sheikh2024impact}, we use w2v2 embeddings of speech segments as input representations.
The message-passing mechanism of GCNs aggregates information across connected nodes, allowing to exploit the similarity of pathological cues across multiple segments of speech as well as to potentially overcome label noise issues arising from segments of PD speakers lacking strong pathological cues.
Experimental results establish the advantages of the proposed approach for PD detection and provide further insights on the model mechanisms.
}}

\section{Background and motivation}
Graphs represent relationships (edges) between objects (nodes) and are extensively used to model data across numerous domains~\cite{kersting16, wu2020comprehensive, ortega2018graph}. 
To learn from such graph-structured data, graph neural networks (GNNs) have been developed. 
Various types of GNNs have emerged, including GCNs~\cite{Kipf16}, graph attention networks~\cite{gat}, and GraphSAGE (SAmple and aggreGatE)~\cite{gsage}. Among these, GCNs are particularly valued for their ability to effectively extract task-relevant information while maintaining relatively low computational complexity, making them a preferred option within the graph community~\cite{wu2019simplifying}. 

In recent years, several works in speech processing have exploited the properties of GCNs. Most GCN-based studies, such as those on emotion recognition~\cite{shirian2021compact, chandola2024serc}, speaker verification~\cite{he2024study}, and anti-spoofing~\cite{chen2023graph}, focus on graph classification problems where a separate graph is created for speech segments based on time frames. 

However, in PD detection where pathological speakers share common attributes related to PD, it is more logical to establish inter-speaker connections by building a graph that connects all speech segments rather than creating a separate graph for each speaker. 
This approach naturally frames the problem as a node classification task, where each node represents a segment of speech. 
To the best of our knowledge, such an approach is the first in the speech processing area that exploits a graph-structure among the speech segments.

\begin{figure}
     \centering
    \begin{subfigure}[b]{0.45\linewidth}
     \includegraphics[scale=0.43]{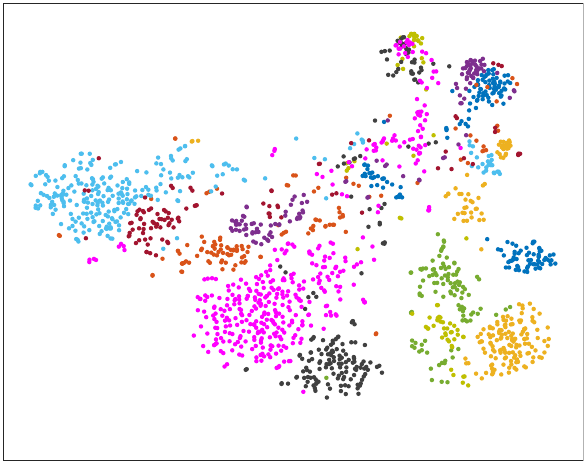}
         \caption{}
        \label{fig:spk_wise}
    \end{subfigure}
    \hfill
    \begin{subfigure}[b]{0.45\linewidth}
    \includegraphics[scale=0.43]{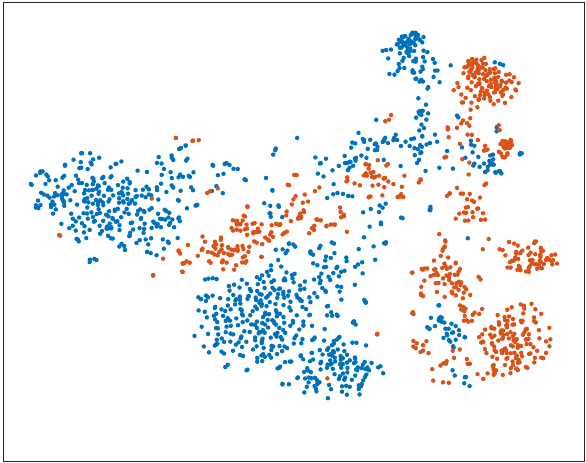}
         \caption{}
        \label{fig:label_wise}
    \end{subfigure}
     \caption{t-SNE (perplexity 100) of the w2v2 embeddings (averaged across time) of segments  from $5$ healthy and $5$ pathological speakers from the PC-GITA database: (a) each color represents a speaker and (b) each color represents the label of the segment, with blue denoting healthy segments and red denoting PD segments. } 
    \label{fig:tsne}
\end{figure}

To further highlight our rationale for using GCNs for PD detection, Fig.~\ref{fig:tsne} presents a t-SNE visualization of the w2v2 embeddings (averaged across time) for speech segments from $5$ healthy speakers and $5$ PD patients from the PC-GITA database~\cite{orozco2014new} (cf. Section~\ref{sec:exp_setup}).
In Fig.~\ref{fig:tsne}a,  the colors represent the different speakers from whom the segments are extracted, while in Fig.~\ref{fig:tsne}b, the colors indicate whether the segments are labeled as healthy (blue) or pathological (red).
The latent space depicted in Fig.~\ref{fig:tsne}b demonstrates that segments sharing the same label tend to cluster closely together, even across different speakers. Furthermore, the overlap between clusters of healthy and PD segments in Fig.~\ref{fig:tsne}(b) indicates that not all segments from a pathological speaker exhibit PD biomarkers.
By clustering segments with shared characteristics
across both speaker and PD groupings, GCNs can effectively diffuse the relevant information.

\section{GCN for PD Detection}

We approach PD detection as a graph node classification problem where some speech segments are labeled (i.e., the training data) and the aim is to predict the labels of the unlabeled segments (i.e., the testing data).
For each available segment $i \in \{1, 2, \cdots, n\}$, with $n$ being the total number of segments in the database, the $m$--dimensional node representation \(\mathbf{x}_i\) is obtained from  the w2v2 embeddings computed as outlined in Section~\ref{sec:exp_setup}.

\subsection{Graph Construction}
\label{gc:graph_cons}
The most straightforward idea for constructing the graph is to calculate the distance between segment representations $\mathbf{x}_i$ and set a threshold below which an edge is established. However, this approach has two key drawbacks. 
First, it relies on a threshold to be tuned which increases the computational cost of the method, particularly since the threshold is a real number. 
Second, this method inadequately controls the local connectivity of the graph. While we can adjust this threshold to achieve a certain number of edges, we lack control over their distribution, potentially resulting in some nodes being overly connected while others remaining isolated. 
To address these issues, we adopt a topological approach as in~\cite{chen2019graph}. 
To this end, we connect each segment to the $k$ segments with the most similar neighbors, where $k$ is an integer hyper-parameter of the approach (cf. Section~\ref{sec:exp_setup}).
This method enables precise local control over graph connectivity, efficiently connecting similar but spatially separated components within the latent space.
To this end, we compute the $(n \times n)$--dimensional kernel $\mathbf{K}^{d}$ where the entry in the $i$-th row and $j$-th column $\mathbf{K}^{d}_{ij}$ represents the similarity between the representations $\mathbf{x}_i$ and $\mathbf{x}_j$.
This similarity is computed as
\begin{align}
   \mathbf{K}^{d}_{ij}  &= \exp({-d(\mathbf{x}_i,\mathbf{x}_j)/h}),
    \label{eq:kernek}
\end{align}
where $d$ is a user-defined distance measure (cf. Section~\ref{sec:exp_setup}) and $h$ is the average distance across all distinct representations pairs, i.e., $h = \frac{2}{n (n-1)} \sum_{i > j} d(\mathbf{x}_i,\mathbf{x}_j)$.
The neighborhood of segment $\mathbf{x}_i$ is represented by the $i$-th column of the kernel and is denoted by $\mathbf{K}^{d}(:,i)$. 
We then compute the Euclidean distance between all column pairs in the kernel \(\mathbf{K}^{d}\). 
A segment \(i\) is connected to a segment \(j\) if \(\mathbf{K}^{d}(:,i)\) is among the top $k$ closest columns to \(\mathbf{K}^{d}(:,j)\), or vice versa, if \(\mathbf{K}^{d}(:,j)\) is among the top $k$ closest columns to \(\mathbf{K}^{d}(:,i)\). 
This procedure results in a symmetric graph structure described by the $(n \times n)$--dimensional adjacency matrix \(\mathbf{A}\), where \(\mathbf{A}_{i,j} = 1\) if segments \(i\) and \(j\) are connected in the graph and \(\mathbf{A}_{i,j} = 0\) otherwise.

\subsection{PD Detection} 
Following the method outlined in Section~\ref{gc:graph_cons}, the entire dataset is transformed into a graph \(\mathbb{G} = (\mathbf{A}, \mathbf{X})\), with $\mathbf{A}$ the previously defined adjacency matrix and \(\mathbf{X}\) the $n \times m$--dimensional matrix of node representations.
We process this graph using a GCN~\cite{Kipf16}, which updates node representations iteratively by propagating information through neighboring nodes, exploiting both the initial representations $\mathbf{X}$ and the structure $\mathbf{A}$.

At each layer $l$ of the GCN with $L$ layers, node representations $\mathbf{X}^{(l)}$ are updated as
\begin{align}
\mathbf{X}^{(l+1)} = \text{ReLU}\left(\tilde{\mathbf{A}} \mathbf{X}^{(l)} \mathbf{W}^{(l)}\right),
\label{eq:GCN}
\end{align} 
with $\mathbf{X}^{(0)}$ being the input node representation $\mathbf{X}$, \( \mathbf{W}^{(l)} \) being the trainable weights, \(\widetilde{\mathbf{A}}\) being a matrix term (derived from the adjacency matrix $\mathbf{A}$) used to stabilize the gradient, and the ReLU non-linearity applied element-wise.\footnote{For additional details, the reader is referred to~\cite{Kipf16}.}

The output representations of the GCN are then processed by a linear layer followed by \emph{softmax} to predict the class label of each segment.  
While the graph is constructed using all segments in the database (i.e., training and test segments) the weights $\mathbf{W}^{(l)}$ and the linear layer are trained for PD detection using only the segments belonging to the training data.

\section{Experimental Settings}
\label{sec:exp_setup}
In the following, we present the various experimental settings used for generating the experimental results in Section~\ref{dicuss}.

\emph{Dataset.} \enspace We use the PC-GITA pathological speech dataset, which contains recordings from gender- and age-balanced groups of $50$ patients diagnosed with PD and $50$ healthy speakers~\cite{orozco-arroyave-etal-2014-new}.
In line with the state-of-the-art literature, we train PD detection models using only controlled speech recordings of $10$ sentences and a phonetically balanced text for each speaker. 
In addition, \cite{sheikh2024impact} has recently shown that w2v2 embeddings can successfully extract pathological cues also from spontaneous speech. Hence, to show the applicability of the proposed model for different speech modes, we also analyze the performance of PD detection models using only spontaneous speech recordings for each speaker.

It should be noted that further analysis on additional databases is required to fully validate the benefits of the proposed model, which remains a topic for future investigation. Currently, the PC-GITA database is the only high-quality open-source database of healthy and PD speech. As demonstrated in~\cite{janbakhshi_ua}, other existing databases, such as UA-Speech~\cite{ua_speech} and TORGO~\cite{rudzicz2012torgo}, are unsuitable for validating pathological speech detection models, as models trained on them may incorrectly interpret recording artifacts as indicators of pathology.

\emph{Input representations.} \enspace 
Prior to computing input representations, available utterances are segmented into $500$ ms segments with a $50$\% overlap.
For a given speech segment, we extract embeddings from the first layer of the transformer module of w2v2 as in~\cite{amiri24_interspeech, sheikh2024impact}.
We use the XLRS53 version of w2v2, which has been pre-trained on 56K hours of unlabeled multilingual audio data~\cite{conneau2020unsupervised}. 
The extracted embeddings are averaged across time to obtain the input representation $\mathbf{x}_i$ for the $i$-th speech segment.

\emph{Baselines.} \enspace To effectively evaluate our proposed GCN-based PD detecion method, we consider two baseline models. 
The first baseline model consists of a fully connected~(FC) layer applied to the input representations.
This model represents the state-of-the-art model when using w2v2 embeddings for PD detection~\cite{amiri24_interspeech, janbakhshi_ua, javanmardi2023wav2vec, sheikh2024impact}.
The second baseline model is a k-nearest neighbors (KNN) approach considered here due to its similarity to our proposed approach (in terms of using a distance measure to compare input representations).
The comparison between the GCN and KNN approaches allows us to assess whether the improvement in PD detection performance achieved through the GCN apporach is merely due to the distance measure or due to the added benefit of using a GCN.
To investigate the impact of the used distance measure for the KNN and GCN approaches, we consider three different distance measures, i.e., the Euclidian distance $d(\mathbf{x}_i,\mathbf{x}_j) = |\!| \mathbf{x}_i -\mathbf{x}_j  |\!|_2 $, the cosine distance $d(\mathbf{x}_i,\mathbf{x}_j) =  1 - \mathbf{x}_i \cdot \mathbf{x}_j / |\!|\mathbf{x}_i |\!|_2|\!|\mathbf{x}_j |\!|_2$, and the Manhattan distance $d(\mathbf{x}_i,\mathbf{x}_j) = |\!| \mathbf{x}_i -\mathbf{x}_j  |\!|_1$.

\emph{Training and Evaluation.} \enspace Evaluation is performed using a 10-fold stratified speaker-independent cross-validation framework. 
In each fold, 80\% of the speakers are used for training, 10\% for validation, and 10\% for testing. 
Performance is evaluated using the speaker-level accuracy computed through soft voting of the softmax output obtained for all segments belonging to a speaker. 
To account for the bias introduced by the exact split of speakers into training, validation, and test sets, PD detection models are trained using $5$ distinct splits. The reported performance represents the mean and standard deviation of the obtained accuracy across these different splits.

For all considered models, results are obtained through hyperparameter tuning on the validation set, with tailored grid searches conducted for each model. For the FC baseline, the hyperparameter search includes the learning rate \textit{lr} $\in$ $\left\{ 10^{-3}, 10^{-2}\right\}$. For the KNN baseline, the hyperparameter search includes the number of neighbors $k \in$ $\left\{ 1, 2, 3, 5, 7, 10 \right\}$. 
    For the proposed GCN approach, the hyperparameter search includes the learning rate \textit{lr} $\in$ $\left\{10^{-3}, 10^{-4} \right\}$, the number of neighbors $k \in$ $\left\{ 1, 2, 3, 5, 7, 10 \right\}$, and the layer depth $L \in \left\{ 2, 3, 4, 5 \right\}$.     
To implement the GCN model, we use the PyTorch Geometric library, which provides an efficient and flexible framework for handling graph data and performing graph-based computations.

\section{Experimental Results}
\label{dicuss}
In the following, the PD detection accuracy of the proposed GCN-based approach is compared to the FC and KNN baselines for the controlled and spontaneous speech modes. 
Additional analysis on the impact of the number of neighbors $k$ and the number of layers $L$ on the performance of the GCN-based approach are also provided.

{\emph{Performance of baselines and proposed approach.}} \enspace
 Table~\ref{tab:xlrs_gcn} presents the performance of the FC, KNN (for various distance measures), and GCN (for various distance measures) approaches for both controlled and spontaneous speech modes.
 For completeness, the optimal number of neighbors $k$ for the KNN and GCN approaches and the optimal layer depth $L$ for the GCN approach (obtained through hyperparameter tuning on the validation set) are also presented.
 It can be observed that the proposed GCN approach considerably outperforms the FC and KNN baselines for both speech modes, independently of the distance measure used.
 This confirms the effectiveness of the GCN method in utilizing message passing to integrate information across speech samples from diverse patients. In terms of distance measures, in the spontaneous speech mode we observe minimal variation in performance of the GCN approach, with the cosine distance yielding the highest accuracy of $92.8$\%. In contrast, the controlled speech mode exhibits more variation, with the Manhattan distance achieving the best performance of $85.8$\%.
Furthermore, Table~\ref{tab:xlrs_gcn} shows that the highest performance for the KNN baseline is achieved when using the cosine distance measure. However independently of the distance measure used, the simple FC baseline yields a better performance than KNN, which indicates that using distance measures in their simplest form is insufficient. To fully exploit distance information, incorporating graph-based computations as in the proposed approach is necessary, as illustrated by the higher performance of GCN.

\begin{table}[t]
  \centering
  \caption{PD detection accuracy (Acc) using the FC baseline, the KNN baseline, and the proposed GCN approach for both controlled (Ctl) and spontaneous (Spt) speech modes. Several distance measures (Dist) are used for the KNN and GCN approaches, i.e., Euclidean distance (E), cosine distance (C), and Manhattan distance (M). The number of neighbors $k$ and layer depth $L$ (when applicable) are obtained through hyper-parameter tuning on the validation set.}
  \label{tab:xlrs_gcn}
    \scalebox{1.13}
   {
  \begin{tabular}{cccccccc}
    \toprule
    Model & Dist & $k$  & $L$  & Acc (Ctl) & \textit{k} & \textit{L}& Acc (Spt) \\

  \midrule
  FC    & -         &  - & - & 79.0 \textcolor{gray}{$\pm$ 2.9}   & - & - & 84.2 \textcolor{gray}{$\pm$ 3.1} \\
    KNN   & E & 5 & - & 72.4 \textcolor{gray}{$\pm$ 1.8} & 5 & - & 82.6 \textcolor{gray}{$\pm$ 1.9}   \\
    KNN   & C    & 5 & - & 76.8 \textcolor{gray}{$\pm$ 1.3}  &  5 & - & 83.8 \textcolor{gray}{$\pm$ 1.8}\\
    KNN   & M  & 1  & - & 73.4 \textcolor{gray}{$\pm$ 2.3} &  1 & - & 80.8 \textcolor{gray}{$\pm$ 1.5} \\
    \midrule
    \midrule
    GCN   & E   & 3 & 3 &82.0 \textcolor{gray}{$\pm$ 2.2} &  5 &  2 & 91.8 \textcolor{gray}{$\pm$ 1.3} \\
    GCN   & C      & 5 & 2 & 82.2 \textcolor{gray}{$\pm$ 2.2} &  3 &  3 & \textbf{92.8} \textcolor{gray}{$\pm$ 2.4}\\
    GCN   & M   & 3 & 3 & \textbf{85.8} \textcolor{gray}{$\pm$ 2.3} &  7 &  3 & 92.0 \textcolor{gray}{$\pm$ 1.2} \\
    \bottomrule
  \end{tabular}
  }
\end{table}

\par
\begin{figure}
 \hspace{-0.2cm}
    \centering 
    \begin{subfigure}[b]{0.45\linewidth}    
        \centering
        \begin{tikzpicture}[scale=0.45]
        \hspace{0cm}
        \begin{axis}[
            xlabel={$k$},
            xlabel style={font=\Huge, xshift=-1pt},
             ylabel={Acc. (\%)},
            ylabel style={font=\Huge, yshift=-5pt}, 
            grid=both,
            xmin=1, xmax=10,
            ymin=70, ymax=100,
            xtick={1,2,3,5,7,10},  
            ticklabel style={font=\LARGE},
            ytick={10,20,30,40,50,60,70,80,90,100},
            ticklabel style={font=\LARGE},
            legend pos=north east,
            legend style={xshift=0.18cm, yshift=0.15cm, font=\Large},
            legend cell align={left},
            ymajorgrids=true,
            grid style=dashed,                 
        ]
        
        \addplot[
            color=blue,
            mark=*,
            mark size=4,
            line width=1.5pt,
            ]
            coordinates {
            (1,73.8)(2,81.6)(3,82.0)(5,80.4)(7,75.8)(10,72.6)
            };

        \addplot[
            color=red,
            mark=square*,
            mark size=4,
            line width=1.5pt,
            ]
            coordinates {
            (1,73.8)(2,80.8)(3,82.2)(5,78.0)(7,75.0)(10,72.2)
            };
        \addplot[
            color=violet,
            mark=triangle*,
            mark size=4,
            line width=1.5pt,
            ]
            coordinates {
            (1,73.8)(2,83.8)(3,85.8)(5,82.0)(7,80.2)(10,74.4)
            };
        
        
        \legend{Euclidean, Cosine, Manhattan}
        \end{axis}
        \end{tikzpicture}
        \caption{}
        \label{fig:neigh_impact_srt}

    \end{subfigure}
         \hfill \hspace{5pt}
    \begin{subfigure}[b]{0.45\linewidth}
        \centering
        \begin{tikzpicture}[scale=0.45]
        \begin{axis}[
             title style={font=\LARGE},
            xlabel={$k$},
            xlabel style={font=\Huge, xshift=-5pt},
             ylabel={Acc. (\%)},
            ylabel style={font=\Huge, yshift=-5pt},
            grid=both,
            xmin=1, xmax=10,
            ymin=70, ymax=100,
            xtick={1,2,3,5,7,10},
            ticklabel style={font=\LARGE},
            ytick={10,20,30,40,50,60,70,80,90,100},
            ticklabel style={font=\LARGE},
            legend pos=south east,
            legend style={xshift=0.18cm, yshift=-0.15cm, font=\Large},
            legend cell align={left},
            ymajorgrids=true,
            grid style=dashed,    
        ]
          
            
        \addplot[
            color=blue,
            mark=*,
            mark size=4,
            line width=1.5pt,
            ]
            coordinates {
            (1,80.4)(2,89.4)(3,92.2)(5,91.8)(7,91.2)(10,89.4)
            };

        \addplot[
            color=red,
            mark=square*,
            mark size=4,
            line width=1.5pt,
            ]
            coordinates {
            (1,81.0)(2,89.0)(3,92.8)(5,88.8)(7,88.0)(10,87.2)
            };
            \addplot[
            color=violet,
            mark=triangle*,
            mark size=4,
            line width=1.5pt,
            ]
            coordinates {
            (1,81.0)(2,91.8)(3,93.0)(5,92.0)(7,92.0)(10,89.0)
            };
        \legend{Euclidean, Cosine, Manhattan}
        \end{axis}
        \end{tikzpicture}
         \caption{}
        \label{fig:neigh_impact_mono}
       
    \end{subfigure}
    \caption{Impact of number of neighbors $k$ on PD detection accuracy of the proposed GCN approach using different distance measures for (a) controlled speech and (b) spontaneous speech.}
    \label{fig:combined_neigh_impact}
\end{figure}
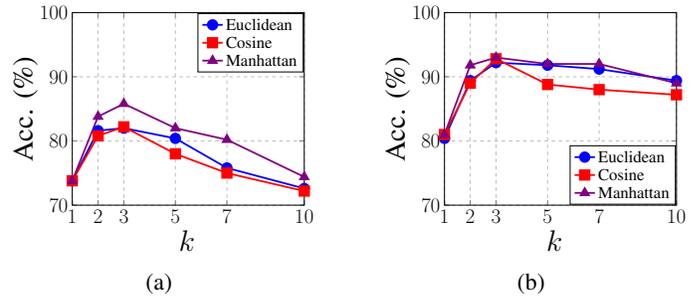

 
\emph{Impact of the number of neighbors $k$ in the GCN approach}. \enspace 
To investigate the impact of the number of neighbors $k$ in the GCN approach, in the following we fix the layer depth $L$ and examine the performance when varying $k$.
For each distance measure and speech mode, the layer depth $L$ is fixed to the values presented in Table~\ref{tab:xlrs_gcn}.
Fig.~\ref{fig:combined_neigh_impact} illustrates the influence of the number of neighbors on the proposed GCN approach for both speech modes.
It can be observed that as the number of neighbors for a particular node increases, the detection performance decreases for all considered distance measures. This decline is higher in the controlled speech mode (Fig. \ref{fig:neigh_impact_srt}) than the spontaneous speech mode (Fig. \ref{fig:neigh_impact_mono}). These results suggest that sparser graphs created from speech segments of various patients tend to yield better PD detection performance.

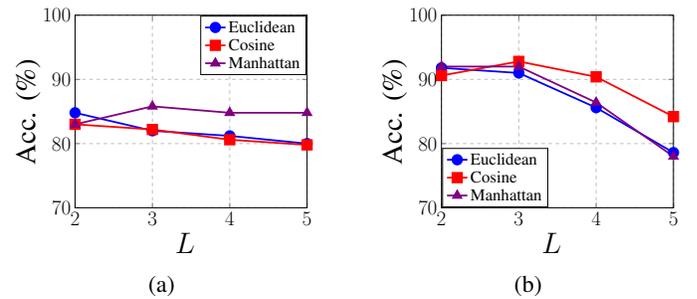
\begin{figure}
     \centering
     \begin{subfigure}[b]{0.45\linewidth}
        \centering      
         \begin{tikzpicture}[scale=0.45]
        \begin{axis}[
            title style={font=\LARGE},
            xlabel={$L$},
            xlabel style={font=\Huge, xshift=-5pt},
             ylabel={Acc. (\%)},
            ylabel style={font=\Huge, yshift=-5pt},
            grid=both,
            xmin=2, xmax=5,
            ymin=70, ymax=100,
            xtick={2,3,4,5},
            ytick={10,20,30,40,50,60,70,80,90,100},
            ticklabel style={font=\LARGE},
            legend pos=north east,
            legend style={xshift=0.18cm, yshift=0.15cm, font=\Large},
            legend cell align={left},
            ymajorgrids=true,
            grid style=dashed,                     
        ]
        \addplot[
            color=blue,
            mark=*,
            mark size=4,
            line width=1.5pt,
            ]
            coordinates {
            (2,84.8)(3,82.0)(4,81.2)(5,80.0)
            };

          \addplot[
            color=red,
            mark=square*,
            mark size=4,
            line width=1.5pt,
            ]
            coordinates {
            (2,83.0)(3,82.2)(4,80.6)(5,79.8)
            };

         \addplot[
            color=violet,
            mark=triangle*,
            mark size=4,
            line width=1.5pt,
            ]
            coordinates {
            (2,83.0)(3,85.8)(4,84.8)(5,84.8)
            };
            
        \legend{Euclidean, Cosine, Manhattan}
        \end{axis}
        \end{tikzpicture}
         \caption{}
        \label{fig:layer_impact_srt}
    \end{subfigure}
    \hfill
    \begin{subfigure}[b]{0.45\linewidth}
        \centering
        \begin{tikzpicture}[scale=0.45]
        \begin{axis}[
            title style={font=\LARGE},
            xlabel={$L$},
            ylabel={Acc. (\%)},
            xlabel style={font=\Huge, xshift=-5pt},
            ylabel style={font=\Huge, yshift=-5pt},
            grid=both,
            xmin=2, xmax=5,
            ymin=70, ymax=100,
            xtick={2,3,4,5},
            ticklabel style={font=\LARGE},
            ytick={10,20,30,40,50,60,70,80,90,100},
            ticklabel style={font=\LARGE},
            legend pos=south west,
            legend style={xshift=-0.18cm, yshift=-0.15cm, font=\Large},
            legend cell align={left},
            ymajorgrids=true,
            grid style=dashed,                     
        ]
        \addplot[
            color=blue,
            mark=*,
            mark size=4,
            line width=1.5pt,
            ]
            coordinates {
            (2,91.8)(3,91.0)(4,85.6)(5,78.6)
            
            };

          \addplot[
            color=red,
            mark=square*,
            mark size=4,
            line width=1.5pt,
            ]
            coordinates {
            (2,90.6)(3,92.8)(4,90.4)(5,84.2)
            };

         \addplot[
            color=violet,
            mark=triangle*,
            mark size=4,
            line width=1.5pt,
            ]
            coordinates {
            (2,92.0)(3,92.0)(4,86.4)(5,78.0)
            };
         \legend{Euclidean, Cosine, Manhattan}   
        \end{axis}
        \end{tikzpicture}
         \caption{}
        \label{fig:layer_impact_mono}
    \end{subfigure}
    
    \caption{Impact of layer depth $L$ on PD detection accuracy of the proposed GCN approach using different distance measures for (a) controlled speech and (b) spontaneous speech. }
    \label{fig:combined_impact}
    \vspace{-10pt}
\end{figure}

\emph{Impact of the layer depth $L$ in the GCN approach}. 
To investigate the impact of the layer depth $L$ in the GCN approach, in the following we fix the number of neighbors $k$ and examine the performance when varying $L$.
For each distance measure and speech mode, $k$ is fixed to the values presented in Table~\ref{tab:xlrs_gcn}.
The results presented in Fig.~\ref{fig:combined_impact} show that there is typically a decline in PD detection performance of the GCN as the layer depth increases, with the decline being more pronounced for the spontaneous speech mode setting in Fig.~\ref{fig:layer_impact_mono}. Such an effect is expected and well-documented in the GCN literature and it arises from saturation that occurs after certain depths in the GCN~\cite{luan19}. 

In summary, the presented results show that a shallow depth of $2$ or $3$ layers and a small neighborhood size (less than $5$) typically ensures an advantageous performance when using a GCN for PD detection. This demonstrates that graph-based approaches are highly effective compared to the state-of-the-art FC layers, with only a modest increase in computational complexity.

\section{Conclusion}
In this paper we have proposed a novel node classification approach for PD detection using GCNs, constructing a graph from speech segments using all available speakers. Experimental results have shown that GCNs considerably improve the PD detection accuracy compared to baseline methods, regardless of the distance measure used to define connections between segments.
Future research will explore incorporating additional meta data into the graph construction process such as gender or age information.

\newpage
\bibliographystyle{IEEEbib}
\bibliography{strings,refs}

\end{document}